\documentclass{article}
\usepackage[skip=10pt]{caption}

\usepackage{spconf,amsmath,graphicx}
\usepackage{hyperref}
\usepackage{url}
\usepackage{booktabs}
\usepackage{bm}
\usepackage{array}
\usepackage{multirow}
\usepackage{units}
\usepackage[acronym]{glossaries}
\usepackage{enumitem}
\usepackage[
backend=biber,
style=ieee,
citestyle=numeric-comp,
maxbibnames=10,
maxcitenames=10,
doi=false,isbn=false,url=false,eprint=false
]{biblatex}

\defbibheading{bibliography}[\refname]{}

\DeclareSourcemap{
    \maps[datatype=bibtex, overwrite=true]{
        \map{
            \step[fieldsource=booktitle,
            match=\regexp{.*Interspeech.*},
            replace={Proc. Interspeech}]
            \step[fieldsource=journal,
            match=\regexp{.*INTERSPEECH.*},
            replace={Proc. Interspeech}]
            \step[fieldsource=booktitle,
            match=\regexp{.*ICASSP.*},
            replace={Proc. ICASSP}]
            \step[fieldsource=booktitle,
            match=\regexp{.*icassp_inpress.*},
            replace={Proc. ICASSP (in press)}]
            \step[fieldsource=booktitle,
            match=\regexp{.*Acoustics,.*Speech.*and.*Signal.*Processing.*},
            replace={Proc. ICASSP}]
            \step[fieldsource=booktitle,
            match=\regexp{.*International.*Conference.*on.*Learning.*Representations.*},
            replace={Proc. ICLR}]
            \step[fieldsource=booktitle,
            match=\regexp{.*International.*Conference.*on.*Computational.*Linguistics.*},
            replace={Proc. COLING}]
            \step[fieldsource=booktitle,
            match=\regexp{.*SIGdial.*Meeting.*on.*Discourse.*and.*Dialogue.*},
            replace={Proc. SIGDIAL}]
            \step[fieldsource=booktitle,
            match=\regexp{.*International.*Conference.*on.*Machine.*Learning.*},
            replace={Proc. ICML}]
            \step[fieldsource=booktitle,
            match=\regexp{.*North.*American.*Chapter.*of.*the.*Association.*for.*Computational.*Linguistics:.*Human.*Language.*Technologies.*},
            replace={Proc. NAACL}]
            \step[fieldsource=booktitle,
            match=\regexp{.*Empirical.*Methods.*in.*Natural.*Language.*Processing.*},
            replace={Proc. EMNLP}]
            \step[fieldsource=booktitle,
            match=\regexp{.*Association.*for.*Computational.*Linguistics.*},
            replace={Proc. ACL}]
            \step[fieldsource=booktitle,
            match=\regexp{.*Automatic.*Speech.*Recognition.*and.*Understanding.*},
            replace={Proc. ASRU}]
            \step[fieldsource=booktitle,
            match=\regexp{.*Spoken.*Language.*Technology.*},
            replace={Proc. SLT}]
            \step[fieldsource=booktitle,
            match=\regexp{.*Speech.*Synthesis.*Workshop.*},
            replace={Proc. SSW}]
            \step[fieldsource=booktitle,
            match=\regexp{.*workshop.*on.*speech.*synthesis.*},
            replace={Proc. SSW}]
            \step[fieldsource=booktitle,
            match=\regexp{.*Advances.*in.*neural.*information.*processing.*},
            replace={Proc. NeurIPS}]
            \step[fieldsource=booktitle,
            match=\regexp{.*Advances.*in.*Neural.*Information.*Processing.*},
            replace={Proc. NeurIPS}]
            \step[fieldsource=booktitle,
            match=\regexp{.*Workshop.*on.* Applications.* of.* Signal.*Processing.*to.*Audio.*and.*Acoustics.*},
            replace={Proc. WASPAA}]
            \step[fieldsource=publisher,
            match=\regexp{.+},
            replace={{}}]
            \step[fieldsource=month,
            match=\regexp{.+},
            replace={{}}]
            \step[fieldsource=location,
            match=\regexp{.+},
            replace={{}}]
            \step[fieldsource=address,
            match=\regexp{.+},
            replace={{}}]
            \step[fieldsource=organization,
            match=\regexp{.+},
            replace={{}}]
        }
    }
}

\title{Augmenting text for spoken language understanding with Large Language Models}
\name{\begin{tabular}[c]{@{}c@{}c@{}c@{}c@{}c@{}} Roshan Sharma$^1$\sthanks{The first author worked on this while at Meta}, Suyoun Kim$^2$, Daniel Lazar$^2$, Trang Le$^2$, Akshat Shrivastava$^2$,  \\ Kwanghoon An$^2$, Piyush Kansal$^2$, Leda Sari$^2$, Ozlem Kalinli$^2$, Michael Seltzer$^2$ \end{tabular}
}
\address{
$^1$Carnegie Mellon University, Pittsburgh, USA and $^2$Meta, Seattle, USA
}

\begin{document}
\vspace{-0.5cm}
\ninept
\maketitle
\begin{abstract}
Spoken semantic parsing (SSP) involves generating machine-comprehensible parses from input speech. Training robust models for existing application domains represented in training data or extending to new domains requires corresponding triplets of speech-transcript-semantic parse data, which is expensive to obtain. In this paper, we address this challenge by examining methods that can use transcript-semantic parse data (unpaired text) without corresponding speech. First, when unpaired text is drawn from existing textual corpora, Joint Audio Text (JAT) and Text-to-Speech (TTS) are compared as ways to generate speech representations for unpaired text. Experiments on the STOP dataset show that unpaired text from existing and new domains improves performance by 2\% and 30\% in absolute Exact Match (EM) respectively. Second, we consider the setting when unpaired text is not available in existing textual corpora. We propose to prompt Large Language Models (LLMs) to generate unpaired text for existing and new domains. Experiments show that examples and words that co-occur with intents can be used to generate unpaired text with Llama 2.0. Using the generated text with JAT and TTS for spoken semantic parsing improves EM on STOP by 1.4\% and 2.6\% absolute for existing and new domains respectively.

\end{abstract}
\section{Introduction}
\label{sec:intro}


Spoken Language Understanding (SLU) is essential for many real-world applications today including conversational agents and virtual assistants. Spoken Semantic Parsing (SSP) is the SLU task that involves transforming a recording to a machine-comprehensible parse tree ~\cite{wang2023treepiece}. End-to-end models ~\cite{arora2023study} operate directly on speech while cascade models~\cite{futami2023pipeline} generate a semantic parse based on the transcript. Two-pass deliberation models ~\cite{le2022deliberation} combine the best of both worlds, by using first-pass transcripts and speech embeddings to improve spoken semantic parsing. However, training such models with supervision requires matched triplets of speech, transcript, and semantic parse. Annotating these triplets is expensive, which limits the size of training data, and consequently model performance.

The need for matched data can be alleviated by developing methods that can use only text data. Text data (transcript-semantic parse) is more easily obtained than speech -- either from existing textual corpora or by prompting Large Language Models (LLMs), and training models with a small amount of paired speech-text data and a large amount of unpaired text is useful.  It is non-trivial to incorporate text-only data into end-to-end models because model outputs cannot be obtained without speech inputs. Prior work has explored the use of text data for speech recognition ~\cite{wang2020,toshniwal2018comparison,hori2019cycle}. External language models trained on text can be used to interpolate token prediction probabilities ~\cite{meng22_interspeech}, but require additional memory, making them unsuitable for on-device applications. Coordinated learning methods ~\cite{chen22r_interspeech,sainath2023joist} project speech and text to a shared embedding space for speech recognition, but such models require significant amounts of paired speech-text data to learn robust mappings. The final class of work generates speech representations for unpaired speech - Joint Audio Text (JAT)~\cite{kim2022joint} uses mean speech embeddings from paired data to represent unpaired text. This is computationally inexpensive, but the speech embeddings do not contain information embedded in real speech. In contrast, synthetic speech from Text-to-speech (TTS) models~\cite{wang2020} produce informative speech representations, but they can be expensive to compute.  There are two cases where additional textual data may be acquired for semantic parsing -- (a) to improve models on existing domains (ED) and (b) to support new domains (ND).  In this paper, we compare JAT and TTS for SSP when unpaired text data is drawn from existing and new domains. 
 
 

\begin{figure}[tp]
    \centering
    \includegraphics[width=7cm]{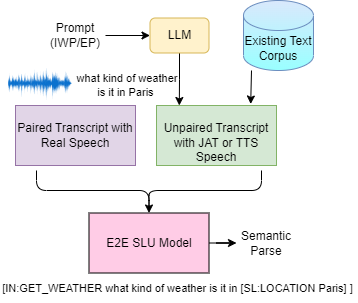}
    \caption{This paper: We describe ways to unpaired text to train deliberation models, where unpaired data can be obtained from LLMs or existing textual corpora. We use JAT or TTS to obtain speech representations of unpaired data }
    \label{fig:overall}
\end{figure}

When unpaired text is not available from existing corpora, we propose prompting Large Language Models (LLMs) ~\cite{ouyang2022training,touvron2023llama,touvron2023llama2} to generate text data for SSP. LLMs are exceptional at generating realistic text based on input prompts, and, in this paper, we use LLama 2.0~\cite{touvron2023llama2} to generate text data. For the ED setup, it is sufficient to generate transcripts since semantic parses can be obtained from transcripts using pre-trained semantic parsers. We describe two prompting methods: (a) intent-word-based prompting (IWP), where the LLM produces transcripts corresponding to a particular intent class and containing words that co-occur with the intent, and (b) exemplar-based prompting (EP), where it generates transcripts that are similar to provided examples. We generate pseudo-labels for the generated utterances using a pre-trained RoBERTa~\cite{liu2020roberta} model and train SSP models using JAT. We find that EP is simpler but IWP generates the desired intent more often. Using data from both methods improves the Exact Match (EM) on STOP data by 1.4 points absolute. 

For the ND setup, pre-trained models for pseudo-labeling are unavailable for the new domain(s), and hence LLMs are used to generate semantic parses directly. The transcript is then inferred from the semantic parse. Exemplar-based prompting (EP) is used with 3 real examples for every possible intent-slot combination to generate large-scale data. We find that the generated data improves EM by 2.3 points absolute over a baseline that uses only 3 examples per combination. 

In summary, this paper makes the following contributions: 
\begin{enumerate}
    \item Extends JAT, previously used for ASR, to end-to-end spoken semantic parsing, and compares JAT with TTS for textual data from existing domains and new domains. 
    \item Develops prompting strategies to generate textual transcripts and semantic parses in existing and new domains using LLMs. 
    \item Demonstrates that LLM-generated textual data can be used in conjunction with JAT and TTS to improve spoken semantic parsing.
\end{enumerate}









\section{Deliberation Model for SLU}
\label{sec:deliberation}
Deliberation-based SLU models~\cite{le2022deliberation, kim2023modality} are two-pass models that predict an ASR transcript in the first pass. Using the first pass transcript and audio, it then generates the semantic parse in the second pass. In contrast to cascade models that utilize separately trained Automatic Speech Recognition (ASR) and SLU components, a deliberation model optimizes both ASR and SLU components jointly. To achieve on-device streaming functionality, the first pass ASR component is implemented using the Recurrent Neural Network Transducer (RNNT) ~\cite{graves2012sequence, kim2021improved, Liu2021MTL}. To maintain transcription accuracy, the ASR component of our deliberation model is trained independently and kept frozen. Our deliberation-based SLU model comprises two primary modules: (1) Fusion, and (2) Decoder. The fusion module combines intermediate audio and text embeddings from the first pass RNNT encoder and predictor respectively. Using Multi-Head Attention~\cite{vaswani2017attention}, the fusion module generates a combined representation that is used by the transformer-based decoder module to predict the target semantic parse sequence.

\section{Speech Representations for Unpaired Text}
\label{sec:available}
\subsection{Joint Audio-Text Training (JAT)}
Joint Audio-Text training (JAT) \cite{kim2022joint} is a recent approach for leveraging unpaired text-only data to improve ASR \cite{ kim2022joint, sainath2023joist, sainath2020attention, wang2020multitask}. Unlike shallow fusion that considers token distributions from an external neural network language model (NNLM), JAT does not require additional model parameters or latency, making it suitable for on-device streaming ASR. The core idea behind JAT is that speech representations for unpaired text can be generated by simply using average speech embeddings computed over available paired speech/text data. In this paper, we use the JAT approach to train our Spoken Language Understanding (SLU) models to enable training with both "speech-text-semantic parse" and "text-semantic parse" datasets.

\subsection{Speech Synthesis with Voicebox}


Voicebox\cite{le2023voicebox} is a state-of-the-art non-autoregressive speech generation model based on Flow Matching \cite{lipman2022flow}. We generate representations for unpaired text by extracting speech features from synthesized speech. Synthetic speech can be obtained by using Voicebox in TTS mode, i.e. where audio is generated by conditioning on input text.
Different from ~\cite{le2023voicebox}, the Voicebox model we use represents input text as graphemes rather than phonemes. To generate audio, we first sample unit durations for each grapheme in the input text using a flow-matching-based duration model and then upsample the grapheme sequence using the unit duration information. This information is used as conditioning to generate the spectrogram using the audio model. Finally, we used a HiFi-GAN ~\cite{kong2020hifi} vocoder to convert the spectrograms into time-domain signals.





 

\section{Generating Textual Data with LLama 2.0}
\label{sec:generating}
LLama 2.0~\cite{touvron2023llama2} is a public open-source large language model trained on large volumes of publicly available data and code with context as large as 4096. In this paper, we use the 13B parameter chat model.

\subsection{Generating Textual Data for Existing Domains}
In the ED setup, we propose to use LLMs to generate transcripts. Corresponding semantic parses are obtained using a pseudo-labeling textual semantic parse model trained on existing paired data. The semantic parse model here takes transcripts as inputs and produces pseudo-label semantic parses as output. Transcripts can be generated using one of two prompting strategies, i.e., intent-word-based or exemplar-based.

\noindent \textbf{Intent Word-based prompting (IWP)}: The goal of IWP is to generate transcripts that may be classified under a certain intent, optionally containing "intent words". Intent words are the words from semantic parses that occur most frequently with given intents after removing stop-words. An example is shown in Figure \ref{fig:intent-word-prompt}. The 40 words that co-occur most frequently with every intent in the STOP data are used as intent words. 40 examples are generated for every intent and intent-word combination. Though IWP produces good synthetic data, it is limited by the fact that words that co-occur less frequently with the intent are less related to the intent. Such examples produced with less relevant intent words may not be classified under the desired intent class. This also limits the amount of synthetic data that can be generated since the LLM cannot generate many unique examples using a small number of intent-intent word combinations. 

\begin{figure}[ht]
\centering
\noindent\fbox{%
\parbox{0.95\linewidth }{%
    \footnotesize 
    \textit{You are working in an intent-and-slot framework where every 
    utterance can be classified under an intent. Here are some examples of intents and a description of their function:}
    
    \textit{1. IN:ADD\_TIME\_TIMER - Creates a new timer} 
    
    \textit{2. IN:GET\_ESTIMATED\_DEPARTURE - gets estimated departure}
    
    \textit{Now, we want to classify intents for the weather application. Given the intents IN:GET\_WEATHER, generate 40 utterances that are classified under this intent.  
    You may use the word "weather" along with names of people and places to generate 40 utterances. 
    Your response should have numbered utterances, with one utterance on each line. Make sure not to repeat any responses. Start with 1.}
}}
\caption{Prompt for IWP-based utterance generation}
\label{fig:intent-word-prompt}
\end{figure}

\begin{figure}[ht]
\centering
\noindent\fbox{%
\parbox{0.95\linewidth }{%
    \footnotesize 
    \textit{Generate 60 more sentences that are similar in intent to the following sentences:}
    
    \textit{1. Is it going to be around 95 in degree Fahrenheit san francisco tomorrow}
    
    \textit{2. Is it around 72 in degree celsius karachi tonight}
    
    \textit{Write one sentence per line. Generate statements and questions with different sentence structure.}
}}
\caption{Prompt for EP-based utterance generation}
\label{fig:exemplar-prompt}
\end{figure}

\begin{figure}[ht]
\centering
\noindent\fbox{%
\parbox{0.95\linewidth }{%
    \footnotesize 
\textit{Each sentence should be enclosed in square brackets [ ]. The first square bracket [ should be followed by an intent that is in uppercase letters and begins with IN:, for example, IN:GET\_WEATHER. Inside the sentence, you should label some nouns with slots, which are also enclosed in brackets [ ]. Slots are in all uppercase letters and begin with SL:, for example, SL:LOCATION. In each sentence, there can only be 1 intent, but there can be many slots. Here are some examples: }

\textit{1. [IN:GET\_WEATHER what kind of weather is in [SL:LOCATION paris ] ]}

\textit{2. [IN:GET\_WEATHER what is the temperature at the [SL:LOCATION north pole ] ]}

\textit{3. [IN:GET\_WEATHER tell me what the weather in [SL:LOCATION central park ] is like ]}

\textit{Please generate more examples with the intent IN:GET\_WEATHER and any of the slots SL:LOCATION. The sentences should have an intent/slot format like [IN:GET\_WEATHER [SL:LOCATION] ], but with some other text, like the examples above.  Write 30 similar sentences and then stop. Use names of people and places in your examples.}
}}
\caption{Prompt for EP-based generation of seqlogical parses}
\label{fig:exemplar_new_domain}
\end{figure}

\noindent \textbf{Exemplar-based Prompting (EP)}: Since LLMs are strong in-context learners~\cite{wei2022emergent}, an alternative approach is to prompt LLMs to generate transcripts based on examples. For every intent-slot combination, we provide up to 4 random example transcripts and ask the model to generate 60 more transcripts that are similar but have diverse sentence structures. An example prompt is shown in Fig \ref{fig:exemplar-prompt}. Though the resulting transcripts may not always correspond to the intent classes from which the examples are drawn, this method enables us to generate larger volumes of data without duplication. 

\noindent \textbf{Semantic Parse generation and Quality Assessment}: Transcripts generated by LLMs are first normalized -- written text is converted to spoken form, punctuation except apostrophes are removed and text is transformed into lower case. Semantic parse pseudo-labels are obtained from these normalized transcripts using a strong RoBERTa-based semantic parser trained on STOP (EM=86.8).  To assess data quality, we compare the intent in the obtained pseudo-labels to the intent in the prompt for IWP or the intent of the provided examples for EP. Intent Match Accuracy (IMA) is defined as the percentage of times the intent of the pseudo-label matches the desired intent of the prompt. 
 
\vspace{-0.4cm}

\subsection{Generating Transcript-Semantic Parse for New Domains}

For new domains, paired data and pre-trained models are not available, and therefore, we would need to directly generate pairs of transcript and semantic parse. One way to do this is to generate pairs of semantic parse and corresponding transcript using LLMs directly, however, maintaining consistency across generated parses and transcripts is challenging for current LLMs. Another alternative is to generate only the seqlogical form of the semantic parse from the LLM and infer the transcript from the parse. The seqlogical form of the parse, unlike the decoupled form, comprises all the words in the transcript along with slot and intent tags. Therefore, the transcript can be obtained from the seqlogical parse merely by removing slot and intent tags. 

\noindent \textbf{Exemplar-based Prompting}: We assume that (a) the intents and slots that must be recognized for the new domain are known, (b) the slots that may occur with every intent, i.e., the intent-slot combinations are known, and (c) some manually annotated examples for every intent-slot combination are known. Using this information, LLMs can be prompted as shown in Figure \ref{fig:exemplar_new_domain} to produce new seqlogical parses for a given intent-slot combinations. The prompt first describes the steps to generate a valid seqlogical parse and then presents up to 3 examples of seqlogical parses with the desired intent-slot combinations. 


\noindent \textbf{Post-processing}: The generated seqlogical parses are checked for invalid placement of brackets, and Out of Vocabulary (OOV) intents and slots. OOV intents were fixed by re-prompting the model to replace OOV intents with correct intents and replace any intents other than the first. Any OOV slots are removed while retaining corresponding slot words.


\section{Experimental Setup}
\label{sec:setup}
\subsection{STOP Data, Model and Metrics}
\textbf{Data}: STOP ~\cite{tomasello2023stop} is a public dataset with real speech for spoken semantic parsing. STOP has data for 8 domains - alarm, event, messaging, music, navigation, reminder, timer, and weather. The data contains 28 unique intents and 82 slot types. 

\noindent \textbf{Metrics}: Exact Match (EM) is used to evaluate all our models. We report EM (No Err) and EM w/ Err, which are the Exact Match accuracies averaged over utterances with no ASR error and averaged over utterances with any ASR error respectively.

\noindent \textbf{Model Configuration}: For the ASR module, we use RNNT with 3 layers of conformer in the encoder, 1 layer of LSTM in the predictor, and 1 linear layer in the joiner. For the deliberation model, we use attention in the Fusion module, 2 transformer encoder layers in the Pooling module, and a transformer decoder layer with a pointer-generator in the Decoder module \cite{kim2023modality}. Models are optimized with Adam~\cite{kingma14_adam}, having a peak learning rate of 8e-3. 

\subsection{Setup: Textual Data from Text Corpora}
For experiments where we assume textual data is available, we split the STOP datasets into two parts. We perform two experiments -- one using the first and second splits as paired and unpaired data respectively and the other using the second and first splits as paired and unpaired data respectively. The average performance across these 2 experiments is reported in each case. In the ED setup, equal amounts of data from every domain are present in the two splits. For the ND setup, STOP is split by domain, where one split contains all training data from 4 domains(messaging, reminder, time, and weather), while the other split contains training data from the other 4 domains (alarm, event, music, and navigation). Both splits are designed to ensure that they have a nearly equal number of utterances. 

\subsection{Setup: Textual Data from LLMs}
When unpaired data is not available, we use Llama 2.0 to generate examples for the ED and ND setups. For the ED setup, LLama 2.0 is used to generate utterances. We then use a pre-trained 12-layer RoBERTa model trained on STOP to generate pseudo-labels for the generated utterances. We augment STOP with the generated LLama 2.0 transcript-semantic parse. JAT is used to represent LLama 2 text. 

For the ND setup, LLama 2.0 generated data is not suitable as a real test set since it does not have matching real speech. Therefore, we choose to partition the existing STOP data into 7 seen domains and 1 new domain - weather. We use exemplar-based prompting to generate transcript-semantic parse pairs for weather. For this, real examples of transcript-semantic parse from STOP are used. We use TTS to generate equivalent speech representations for the generated data. We compare the performance on the weather domain for models trained on (a) 7 domains of STOP, (b) 7 domains of STOP with examples for the weather (with TTS for examples and real speech for 7 domains), (c) 7 domains of STOP with examples and Llama 2.0 generated data, and (d) the topline that uses 7 domains of STOP with real data and TTS.



\section{Experiments}
\label{sec:results}
\subsection{When textual data is available}

Table \ref{tab:same_domain_all} compares the performance of different models for the ED and ND settings where unpaired text is drawn from existing domains and new domains respectively. Across both ED and ND setups, we find that the use of unpaired text improves EM scores. 

\begin{figure}
    \centering
    \includegraphics[width=6.5cm]{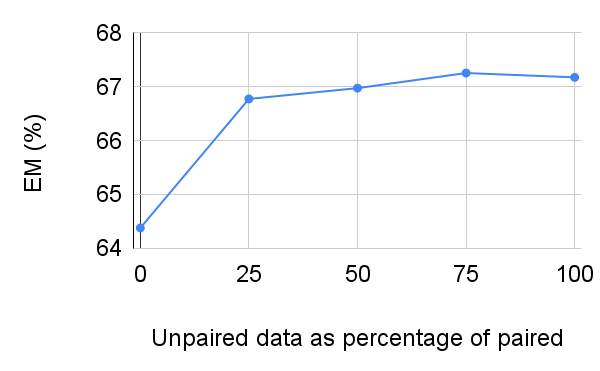}
    \caption{Impact of increasing unpaired text on EM}
    \label{fig:unpaired_data_impact}
\end{figure}

For the ED setup, we find that JAT and TTS achieve similar Exact Match scores. Since JAT is comparable in performance to TTS and relatively inexpensive compared to complex TTS models like Voicebox, JAT is optimal for the ED setup.  Further, the difference between JAT and TTS appears to be primarily on utterances with ASR errors, since synthetic speech representations can be used to reduce the impact of ASR errors on semantic parsing. For the ND setup, we find that though JAT outperforms the baseline, TTS outperforms JAT. This is because new domains may have different entities and domain-specific terms that may be harder to recognize, and TTS provides valid speech representations that can be used to improve predictions based on the first-pass ASR.  Figure \ref{fig:unpaired_data_impact} shows that the amount of unpaired textual data is increased with constant paired data, relative gains increase to a point and saturate.  

\begin{table}[ht]
\caption{Comparing JAT and TTS as speech representations for unpaired text from ED and ND. Number of paired and unpaired utterances, and Exact Match (EM) is reported}
\centering
\resizebox{0.9\columnwidth}{!}{
\begin{tabular}{rl r r r r}
\toprule 
& Model            & \#Pair/\#Unpair     & EM & EM(No Err) & EM w/ Err \\
\midrule 
\parbox[t]{3mm}{\multirow{3}{*}{\rotatebox[origin=c]{90}{\emph{ED}}}} & Baseline &	60.4k / 0 &   64.25 		 &        80.51      &  24.37            \\
& w/ JAT & 60.4k / 60.4k    & 66.92   &   83.90           &   25.25           \\
& w/ TTS & 60.4 / 60.4k  &  \textbf{67.05}  &    \textbf{83.88}          &   \textbf{25.80}          \\ \midrule
\parbox[t]{4mm}{\multirow{3}{*}{\rotatebox[origin=c]{90}{\emph{ND}}}} & Baseline &   
60.7k / 0 & 33.28 &	41.32 &	13.54 \\
& w/ JAT & 60.7k / 60.1k     & 57.74  &	73.34 &	19.50 \\
& w/ TTS & 60.7k / 60.1k &  \textbf{63.95} &	\textbf{80.70} &	\textbf{22.88} \\ \midrule
& Topline & 120.9k / 0 & 67.67   &    84.52           &    26.34          \\
\bottomrule
\end{tabular}
}
\label{tab:same_domain_all}
\end{table}





\subsection{LLama 2.0 Generated Data: ED Setup}

\begin{table}[ht]
\caption{Assessing the impact of augmenting the training data with LLama 2.0 generated utterances and RoBERTa pseudo-labels.EM is Exact Match Accuracy}
\resizebox{1\columnwidth}{!}{
\begin{tabular}{l rrrrr}
\toprule 
Model            & \#Utts & IMA    & EM & EM(No Err) & EM w/ Err \\
\midrule 
STOP Baseline & 160k  & - & 67.37			 &        84.52      &  26.34            \\ 
+ IWP-JAT & 230k & \textbf{68.87}  & 68.12   &   84.96           &   26.82           \\ 
+ EP-JAT & 218k & 64.24 &  68.21  & 85.01	             &       \textbf{27.04}      \\ 
+ (IWP+EP)-JAT & 298k & 67.87 & \textbf{68.75} & \textbf{85.82} & 26.86 \\ \bottomrule
\end{tabular}
}
\label{tab:same_domain_llama}
\end{table}

Table \ref{tab:same_domain_llama} compares various prompting strategies for generating utterances in the same domain using Llama 2.0. We find that combining LLama-generated data with existing STOP data can improve performance across test examples with and without ASR errors. On further analysis, we find that significant improvements are observed across domains with relatively poor performance in the STOP baseline. Between IWP and EP, we find that EP is slightly better. Since EP is not constrained to generate utterances that may be classified under a given intent, the Intent Match Accuracy (IMA) is lower than that of IWP. Combining the data generated from both these strategies further improves performance over the STOP baseline. 

\subsection{LLama 2.0 Generated Data: ND Setup}

\begin{table}[ht]
\caption{Using TTS to generate speech for LLama 2.0 text when unpaired text is in an unseen new domain}
\centering
\resizebox{\columnwidth}{!}{
\begin{tabular}{l r r r}
\toprule 
Model       & \#Utts(Weather)          & Weather EM & Overall EM \\
\toprule 
STOP 7 domain & 0 & 0 &	54.61 \\
+ 3 Examples-TTS &  360  & 48.18  & 61.80 \\
+ Exemplar LLama2-TTS & 2,910  &  \textbf{50.82} &	\textbf{62.29} \\ \midrule
Topline: STOP Weather-TTS & 2,910 &  63.80 & 66.33	 \\  \bottomrule
\end{tabular}
}
\label{tab:new_domain_llama}
\end{table}

Table \ref{tab:new_domain_llama} compares the performance of baseline models that have no data for weather or 360 examples for weather with models that use LLama 2.0 generated data. Llama 2 generated text can improve performance by over 2 points absolute EM but lags behind the performance of a topline that uses data from STOP.

\section{Conclusion}
\label{sec:conclusion}
We address the high cost of manually labeling speech-transcript-semantic parse data for spoken semantic parsing by enabling models to use text-only data. JAT is preferred for unpaired text in existing domains for its efficiency and gain of 2.5 \% EM over a paired data baseline while remaining within 0.1 \% EM of the more computationally expensive TTS.  For unpaired text in new domains, TTS outperforms JAT  by 6 \% absolute EM overall, with a gain of 30.6 \% EM over a paired baseline. 
When text data cannot be obtained from existing text corpora, we propose to prompt LLMs to generate transcript-semantic parse pairs. We show that using different prompting strategies, we can generate unpaired text data in relatively large volumes. Using JAT and TTS, we can leverage this LLM-generated data to further improve SSP by 1.4 \% EM and 2.6 \% EM absolute for existing and new domains.



\vfill\pagebreak


\section{References}
{
\printbibliography
}
\end{document}